\documentclass[final,journal,twocolumn]{IEEEtran}
\pdfoutput=1
\usepackage{graphicx}
\usepackage[cmex10]{amsmath}
\usepackage{amssymb,amsthm}
\usepackage{array}
\usepackage{booktabs}
\usepackage{adjustbox}
\usepackage{xcolor}
\usepackage{cite}
\usepackage{pifont}
\usepackage[normalem]{ulem}
\usepackage[hidelinks=true]{hyperref}

\def\cmark{\ding{51}}
\def\xmark{\ding{55}}

\graphicspath{{./}}
\DeclareGraphicsExtensions{.pdf,.jpg,.png}

\newcommand{\argmax}{\operatornamewithlimits{argmax}}
\DeclareMathOperator*{\maximize}{maximize}
\DeclareMathOperator*{\minimize}{minimize}

\begin{document}

\title{Fine-Grained Object Recognition and Zero-Shot Learning in Remote Sensing Imagery%
    \thanks{Manuscript received June 23, 2017; revised August 10, 2017.
    This work was supported in part by the TUBITAK Grant 116E445
    and in part by the BAGEP Award of the Science Academy.}%
}
\author{%
    Gencer~Sumbul,
    Ramazan~Gokberk~Cinbis,
    and~Selim~Aksoy,~\IEEEmembership{Senior Member,~IEEE}%
    \thanks{The authors are with the Department of Computer Engineering,
    Bilkent University, Ankara, 06800, Turkey. After this work has been done,
    the affliation of Gokberk Cinbis has changed to METU.
    Email: \mbox{gencer.sumbul@bilkent.edu.tr}, \mbox{gcinbis@ceng.metu.edu.tr},
    \mbox{saksoy@cs.bilkent.edu.tr}.}
}

\maketitle

\begin{abstract}
Fine-grained object recognition that aims to identify the type of an object
among a large number of sub-categories is an emerging application with the
increasing resolution that exposes new details in image data. Traditional fully
supervised algorithms fail to handle this problem where there is low
between-class variance and high within-class variance for the classes of
interest with small sample sizes. We study an even more extreme scenario named
zero-shot learning (ZSL) in which no training example exists for some of the
classes. ZSL aims to build a recognition model for new unseen categories by
relating them to seen classes that were previously learned. We establish this
relation by learning a compatibility function between image features extracted
via a convolutional neural network and auxiliary information that describes the
semantics of the classes of interest by using training samples from the seen
classes. Then, we show how knowledge transfer can be performed for the unseen
classes by maximizing this function during inference. We introduce a new data
set that contains $40$ different types of street trees in $1$-foot spatial
resolution aerial data, and evaluate the performance of this model with
manually annotated attributes, a natural language model, and a scientific
taxonomy as auxiliary information. The experiments show that the proposed model
achieves $14.3\%$ recognition accuracy for the classes with no training
examples, which is significantly better than random guess accuracy of $6.3\%$
for $16$ test classes, and three other ZSL algorithms.
\end{abstract}


\section{Introduction}
\label{sec:Introduction}

Advances in sensor technology have increased both the spatial and the spectral
resolution of remotely sensed images. Consequently, the increased resolution
has exposed new details, and has enabled new object classes to be detected and
recognized in aerial and satellite images.

Automatic object recognition has been one of the most popular problems in
remote sensing image analysis where the algorithms aim to map visual
characteristics observed in image data to object classes.
Both the traditional methods that use
various hand-crafted features with classifiers such as support vector
machines and random forests, and the more recent approaches that use deep
neural networks that aim to learn both the features and the classification
rules have been shown to achieve remarkable performance in data sets acquired
from different sources \cite{Hu:2015,Xia:2017}. A common characteristic of such
data sets in the remote sensing literature is that they contain relatively
distinctive classes, with a balanced mixture of urban, rural, agricultural,
coastal, etc., land cover/use classes and object categories, for which
sufficient training data to formulate a supervised learning task are often
available. For example, commonly used benchmark data sets (e.g., UC Merced
and AID \cite{Xia:2017}) pose the classification
problem as the assignment of a test image patch to the most relevant category
among the candidates such as agricultural, beach, forest, freeway, golf course,
harbor, parking lot, residential, and river.
Such data sets have been beneficial in advancing the state-of-the-art by
enabling objective comparisons of different approaches. However, the unconstrained
variety of remotely sensed imagery still leads to many open problems.

\begin{figure}
\centering
\includegraphics[width=\linewidth]{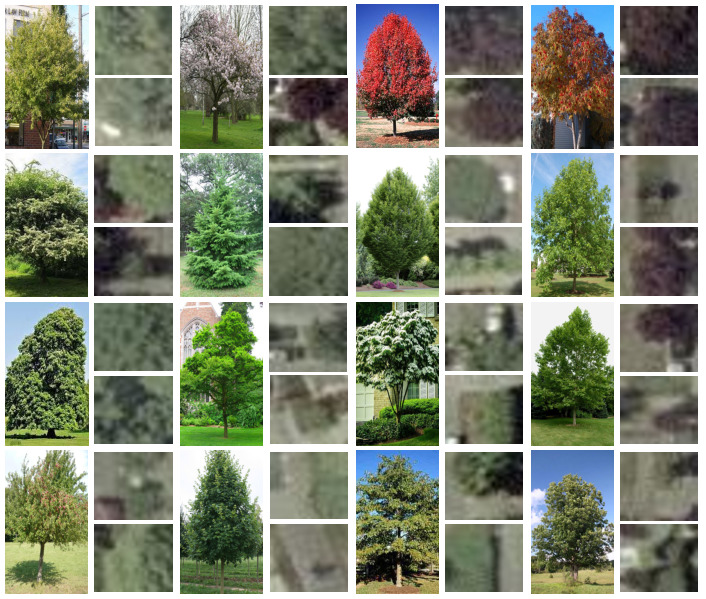}
\caption{Example instances for $16$ classes from the fine-grained tree data set
used in this paper. For each class, a ground-view photograph and two $25 \times 25$
pixel patches from aerial imagery with $1$-foot spatial resolution are shown.
From left to right and top to bottom: \textit{London Plane, Callery Pear, Horse
Chestnut, Common Hawthorn, European Hornbeam, Sycamore Maple, Pacific Maple,
Mountain Ash, Green Ash, Kousa Dogwood, Autumn Cherry, Douglas Fir, Orchard
Apple, Apple Serviceberry, Scarlet Oak, Japanese Snowbell}.}
\label{fig:DataSet}
\end{figure}

A particular problem that has not received any attention in the remote sensing
literature is \emph{fine-grained object recognition} where one is interested in the
identification of the type of an object among a large number of sub-categories.
Figure \ref{fig:DataSet} shows examples from the tree data set used in
this paper. As seen from the $16$ test classes among $40$
types of street trees included in this data set, differentiating the sub-category
can be a very difficult task even when very high spatial resolution image data
are used. We envision that the fine-grained object recognition task will gain
importance in the coming years as both the diversity and the subtleness of
target object classes increase with the constantly improving spatial and
spectral resolution. However, it is currently not clear how the existing
classification models will behave for such recognition tasks.

Fine-grained object recognition differs from other classification and recognition
tasks with respect to two important aspects: small sample size and class
imbalance. Remote sensing has traditionally enjoyed abundance of data, but
obtaining label information has always been an important bottleneck in
classification studies. Common attempts for reducing the effects of limited
training data include regularization in parameter estimation
\cite{Kuo:2002} and feature extraction \cite{Kuo:2007} as well as using
classifier ensembles and utilizing spatial contextual information \cite{Li:2015}.
However, significantly low between-class variance and high within-class
variance in fine-grained recognition tasks limit the use of such statistical
solutions. Another approach for tackling the insufficiency of annotated
samples is to use active learning for interactively collecting new
examples \cite{Tuia:2009,Demir:2011}. However, collecting examples for a very large
number of very similar object categories in fine-grained recognition by using
visual inspection of image data can be very difficult even for domain experts,
as can be seen in the aerial-view examples in Figure \ref{fig:DataSet}.
Furthermore, the acquisition costs for spatially distributed data can make
sample collection via site visits practically unfeasible when one needs to
travel unpredictably long distances to find sufficient number of examples
\cite{Loi:2009}. Class imbalance in training data can also cause problems
during supervised learning, particularly when the label frequencies observed in
training data do not necessarily reflect the distribution of the labels among
future unseen test instances.

Besides these problems, an even more extreme scenario is the \emph{zero-shot
learning} task where \emph{no training examples} exists for some of the classes.
To the best of our knowledge, zero-shot learning for fine-grained object
recognition has not been studied in the remote sensing literature even though
it is a highly probable scenario where new object categories can be introduced
after the training phase or when no training examples exists for several rare
classes that are still of interest.

Zero-shot learning aims to build a recognition model for new categories that
have no training examples by relating them to categories that were previously
learned \cite{Romera-Paredes:2015}.
It is different from the domain adaptation and supervised transfer learning tasks
\cite{Tuia:2016} where at least some training examples are available for the
target classes or the same classes exist in the target domain.
Since no training instances are available for the test categories in zero-shot
learning, image data alone are not sufficient to form the association between
the \emph{unseen} and \emph{seen} classes. Thus, we need to find new sources of
auxiliary information that can act as an intermediate layer for building this
association. Attributes \cite{Ferrari:2007,Farhadi:2009} have been the most popular
source of auxiliary information in the computer vision literature where
zero-shot learning has recently become a popular problem \cite{Lampert:2014}.
Attributes often refer to well-known common characteristics of objects, and can
be acquired by human annotation. They have been successfully used in zero-shot
classification tasks for the identification of different bird or dog species or
indoor and outdoor scene categories in computer vision \cite{Lampert:2014}.
An important requirement in the design of the attributes
is that the required human effort should be small because otherwise resorting
to supervised or semi-supervised learning algorithms by collecting training
samples can be a viable alternative.
An alternative is to use automatic processing of other modalities such as text
documents \cite{Akata:2015}. As the only example in the
remote sensing literature, the Word2Vec model \cite{Mikolov:2013}
that was learned from text documents in Wikipedia was used for zero-shot scene
classification by selecting some of the scene classes in the UC Merced data set
as unseen categories. New relevant attributes that exploit the peculiarities of
overhead imagery should be designed for target object categories of interest
in remotely sensed data sets.

Our main contributions in this paper are as follows.
First, to the best of our knowledge, we present the first study on fine-grained
object recognition with zero-shot learning in remotely sensed imagery. The
proposed approach uses a bilinear function that models the compatibility
between the visual characteristics observed in the input image data and the
auxiliary information that describes the semantics of the classes of interest.
The image content is modeled by features extracted using a convolutional
neural network that is learned from the seen classes in the training data. The
auxiliary information is gathered from three complementary domains: manually
annotated attributes that reflect the domain expertise, a natural language model
trained over large text corpora, and a hierarchical representation of
scientific taxonomy. When the between-class variance is low and the
within-class variance is high, a single source of information is often not
sufficient. Thus, we exploit different representations and comparatively
evaluate their effectiveness. Second, we show how the compatibility function
can be estimated from the seen classes by using the maximum likelihood
principle during the learning phase, and how knowledge transfer can be
performed for the unseen classes by maximizing this function
during the inference phase. Third, we present a new data set that contains $40$
different types of trees with $1$ foot spatial resolution RGB data and
point-based ground truth. We illustrate the use of this data set in a
zero-shot learning scenario by sparing some classes as unseen but it can
also be used in other novel fine-grained object recognition tasks. Fourth, we
present a realistic performance evaluation in a challenging setup by using
different partitionings of the data, making sure that the zero-shot (unseen)
categories are well-isolated from the rest of the classes during both learning
and parameter tuning \cite{Xian:2017}.

The rest of the paper is organized as follows. Section \ref{sec:DataSet}
introduces the fine-grained data set. Section \ref{sec:Methodology} describes
the details of the methodology. Section \ref{sec:Experiments} presents the
experiments. Section \ref{sec:Conclusions} provides the conclusions.

\section{Data set}
\label{sec:DataSet}

There is currently no publicly available remote sensing data set that contains
a large number of classes with high within-class and low between-class variance.
Thus, we created a new data set\footnote{Available at
\url{http://www.cs.bilkent.edu.tr/~saksoy/publications.html}.}
that provides a challenging test bed for fine-grained object recognition research.
We have gathered the data set from two main sources. The first
part corresponds to point GIS data for street trees provided by the Seattle
Department of Transportation in Washington State, USA \cite{trees_data_set}. In
addition to location information in terms of latitude and longitude, the GIS
data contain the scientific name and the common name for each tree. The second
part was obtained from the Washington State Geospatial Data Archive's Puget
Sound orthophotography collection \cite{seattle_orthophoto}. This part
corresponds to $1$ foot spatial resolution aerial RGB images that we mosaiced
over the area covered by the GIS data.

Among the total of $126,\!149$ samples provided for $674$ tree categories,
we chose the top $40$ categories that contain the highest number of instances.
We also carefully went through every single one of the samples, and made sure
that the provided coordinate actually coincides with a tree. Some samples had
to be removed during this process due to mismatches with the aerial data,
probably because of seasonal and temporal differences between ground truth
collection and aerial data acquisition. Finally, each tree is represented as a
$25 \times 25$ pixel patch that is centered at the point ground truth
coordinate where the patch size was chosen as $25$ to cover the largest tree.
Overall, the resulting data set contains a total of $48,\!063$ trees from $40$
different categories. The list of these categories along with the number of
instances in each category is given in Table \ref{data_partition_table}.
We use different splits of this imbalanced data set for a fair and objective
evaluation of fine-grained object recognition with zero-shot learning as
suggested in \cite{Xian:2017} and presented in Section \ref{sec:Experiments}.
Figure \ref{fig:DataSet} illustrates the $16$ categories that are used as the
unseen classes.

\section{Methodology}
\label{sec:Methodology}

In this section, we describe the mathematical formulation of our zero-shot
learning (ZSL) approach and the image and class representations that we utilize
for describing the aerial objects and fine-grained object classes.

\subsection{Zero-shot learning model}
\label{sec:compatibility}

Our goal is to learn a discriminator function that maps a given image $x \in
\mathcal{X}$ to one of the target classes $y \in \mathcal{Y}$ where $\mathcal{X}$
is the space of all images and $\mathcal{Y}$ is the set of all object classes.
By definition of
zero-shot learning, training examples are available only for a subset of the
classes, $\mathcal{Y}_\text{tr} \subset \mathcal{Y}$, which are called the {\em
seen classes}. Therefore, it is not possible to directly use traditional
supervised methods, like decision trees, to build a model that can recognize
the {\em unseen classes}, $\mathcal{Y}_\text{te} \subset \mathcal{Y}$, i.e.,~those
with no training samples, when $\mathcal{Y}_\text{tr} \cap \mathcal{Y}_\text{te}
= \emptyset$.

To overcome this difficulty, we first assume that a vector-space representation,
called {\em class embedding}, is available for each
class. Each class embedding vector is expected to depict (visual) characteristics
of the class such that classification knowledge can be transferred from seen to
unseen classes.

To carry out this knowledge transfer, we utilize a compatibility function $F : \mathcal{X} \times \mathcal{Y} \rightarrow
\mathbb{R}$, which is a mapping from a given image-class pair ($x$,$y$) to a scalar value. This value represents the confidence
in assigning the image $x$ to class $y$.

\begin{figure}
\centering
\includegraphics[width=0.90\linewidth]{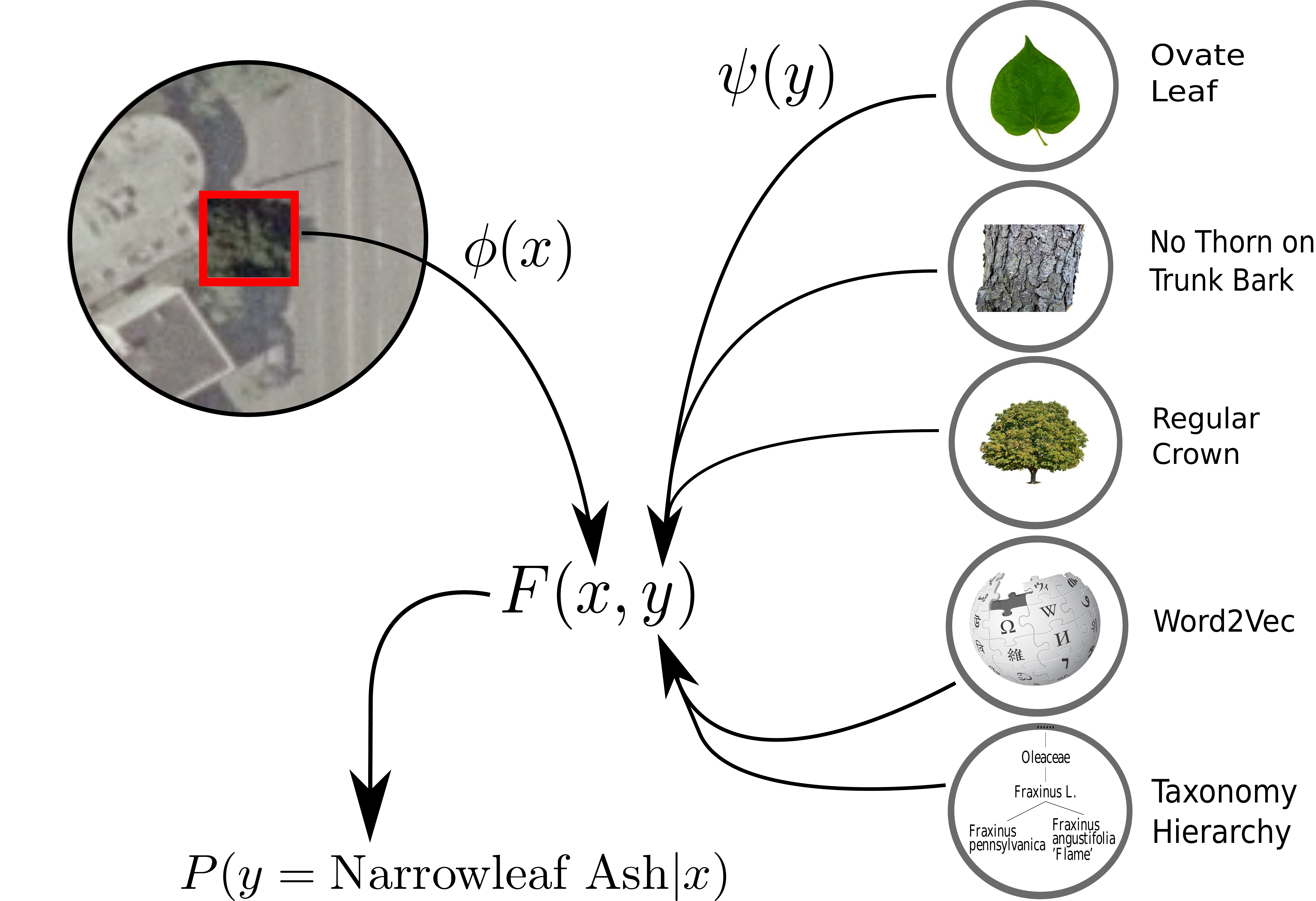}
\caption{Our proposed framework learns the compatibility function $F(x,y)$
between image embedding $\phi(x)$ and class embeddings $\psi(y)$ based on
attributes, word-embeddings from a natural language model, and a hierarchical
scientific taxonomy. The learned compatibility function is then used in recognizing
instances of unseen classes by leveraging their class embedding vectors.}
\label{framework}
\end{figure}

Since examples only from the seen classes are available for learning the
compatibility function, which will be utilized for recognizing instances of the
unseen classes, $F(x,y)$ should employ a class-agnostic model. For this purpose,
following the recent work on ZSL~\cite{Xian:2017}, we define the compatibility
function in a bilinear form, as follows:
\begin{equation}
    F(x,y) = \phi(x)^\top W \psi(y).
    \label{compatibility_func}
\end{equation}
In this equation, $\phi(x)$ is a $d$-dimensional image representation, called
image embedding,
$\psi(y)$ is an $m$-dimensional class embedding vector, and $W$ is a $d \times m$
matrix. This compatibility function can be considered as a class-agnostic model of
a cross-domain relationship between the image representations and class embeddings.
See Figure~\ref{framework} for an illustration.

A number of empirical loss minimization schemes have been proposed for learning such ZSL compatibility functions in recent years.
A detailed evaluation of these schemes can be found in~\cite{Xian:2017}.
In our preliminary experiments, we have investigated the state-of-the-art approaches
of \cite{Akata:2015} and \cite{Romera-Paredes:2015}, and
observed that an intuitive alternative formulation based on an adaptation of multi-class
logistic regression classifier yields comparable to or better results than the others. In our approach,
we define the class posterior probability distribution as the {\em softmax} of compatibility scores:
\def\yp{y^\prime}
\begin{equation}
p(y|x)=\frac{\exp\left(F(x,y)\right)}{\sum_{\yp \in \mathcal{Y}_\text{tr}} \exp\left(F(x,\yp)\right)}
\label{softmax}
\end{equation}
where $\mathcal{Y}_\text{tr} \subset \mathcal{Y}$ is the set of seen (training) classes.
Then, given $N_\text{tr}$ training examples,
we aim to learn $F(x,y)$ using the maximum likelihood principle. Assuming that the data set contains
independent and identically distributed samples, the label likelihood is given by
\begin{equation}
    \maximize_{W \in \mathbb{R}^{d \times m}}
    \prod_{i=1}^{N_\text{tr}} p(y_i|x_i).
\label{objective_func_org}
\end{equation}
The optimization problem can be interpreted as finding the $W$ matrix
that maximizes the predicted true class probabilities of training examples, on average.
Equivalently, the parameters can be found by minimizing the negative log-likelihood:
\begin{equation}
    \minimize_{W \in \mathbb{R}^{d \times m}}
    \sum_{i=1}^{N_\text{tr}} -\log p(y_i|x_i).
\label{objective_func_log}
\end{equation}
To find a local optimum solution, we use stochastic gradient descent (SGD) based optimization. The main idea in SGD
is to iteratively sample a batch of training examples, compute approximate gradient over the batch, and
update the model parameters using the approximate gradient.  In our case, at SGD iteration $t$, the gradient
matrix $G_t$ over a batch $B_t$ of training examples can be computed as follows:
\begin{align*}
    G_t = -\sum_{i \in B_t} \nabla_{W}  \log p(y_i|x_i)
\end{align*}
where the gradient of the log-likelihood term for the $i$-th sample is given by
\begin{align*}
\nabla_{W} \log p(y_i|x_i)
               &= \phi(x_i) \psi(y_i)^\top - \sum_{y \in \mathcal{Y}_\text{tr}} p(y|x_i) \phi(x_i) \psi(y)^\top.
\end{align*}
Given the approximate gradient, the plain SGD algorithm works by subtracting a matrix proportional to
$G_t$, from the model parameters:
\begin{equation}
    W_{t} \leftarrow W_{t-1} - \alpha G_t
\end{equation}
where $W_{t}$ denotes the updated model parameters, and the {\em learning rate} $\alpha$
determines the rate of updates over the SGD iterations. It is often observed
that the learning rate needs to be tuned carefully in order to avoid too large or too small parameter updates, which is
necessary to maintain a stable and steady progress over the iterations.  However, not only finding the right learning rate
is an uneasy task, but also the optimal rate may vary across dimensions and over the
iterations~\cite{goodfellow16book}.

In order to minimize the manual effort for finding a well-performing learning rate policy,
we resort to adaptive learning rate techniques. In particular, we utilize the {\em Adam} technique~\cite{Adam:2014}, which
estimates the learning rate for each model parameter based on the first and second moment estimates of the gradient
matrix. For this purpose, we calculate the running averages of the moments at each iteration:
\begin{align*}
    M_t &= \beta_1 M_{t-1} + (1 - \beta_1) G_t \\
    V_t &= \beta_2 V_{t-1} + (1 - \beta_2) G_t^2
\end{align*}
where $M_t$ and $V_t$ are the first and second moment estimates, $\beta_1$ and $\beta_2$ are the corresponding exponential
decay rates, and $G_t^2$ is the element-wise square of $G_t$. Then, the SGD update step is modified as follows:
\begin{align*}
    W_{t} \leftarrow W_{t-1} - \alpha~\hat{M}_t / (\sqrt{\hat{V}_t}+\epsilon)
\end{align*}
where $\hat{M}_t=M_t/(1-\beta^t_1)$ and $\hat{V}_t=V_t/(1-\beta^t_2)$ are the bias-corrected first and second moment estimates.
These estimates remove the inherent bias towards zero due to zero-initialization of $M_t$ and $V_t$ at $t=0$, which is
particularly important in early iterations. Overall, $\hat{M}_t$ provides a momentum-based approximation to the true gradient
based on the approximate gradients over batches, and $\hat{V}_t$ provides a per-dimension learning rate adaptation based on an
approximation to diagonal Fisher information matrix.

Finally, we should also note that we do not use an explicit regularization term
on $W$ in our training formulation.  Instead, we use {\em early stopping}
as a regularizer. For this, we track the performance of the ZSL model on an independent validation set over
optimization steps, and choose the best performing iteration. Additional
details are provided in Section \ref{sec:Experiments}.

Once the compatibility function (i.e., the $W$ matrix) is learned, zero-shot
recognition of unseen test classes is achieved by assigning the input image
to the class $y^*$ whose vector-space embedding yields the highest score as
\begin{equation}
    y^* = \argmax_{y \in \mathcal{Y}_\text{te}} F(x,y).
\end{equation}

In the next two sections, we explain the details of our image representation and class embeddings, which have central importance in
ZSL performance.

\subsection{Image embedding}
\label{sec:image_embedding}

We employ a deep convolutional neural network (CNN) to learn and extract region
representations for aerial images. The motivation for using a CNN is to be able
to exploit both the pixel-based spectral information and the spatial texture
content. Spectral information available in the three visible bands is not
expected to be sufficiently discriminative for fine-grained object recognition,
and the learned texture representations are empirically found to be superior to
hand-crafted filters.

\begin{figure*}
\centering
\includegraphics[width=\linewidth]{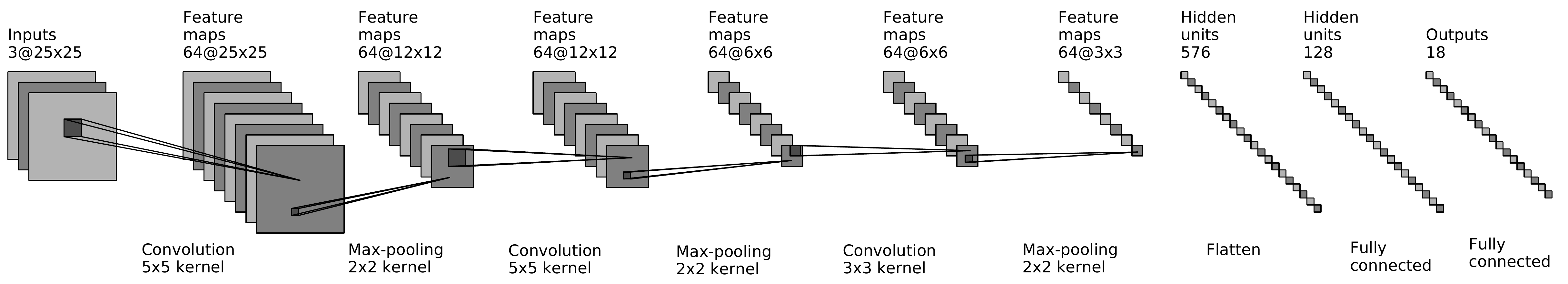}
\caption{Proposed deep convolutional neural network architecture with three
convolutional layers containing $64$ filters each with sizes $5 \times 5$,
$5 \times 5$, and $3 \times 3$, respectively, followed by two fully-connected
layers containing $128$ and $18$ neurons, respectively. We apply max-pooling
after each convolutional layer. The feature map sizes are stated at the top of
each layer.}
\label{fig:CNN}
\end{figure*}

For this purpose, based on our preliminary experiments using only the $18$ seen
classes from our data set (see Section \ref{sec:expsetup} for data set details),
we have developed an architecture that contains three convolutional layers with $5 \times 5$, $5 \times 5$, and $3 \times 3$ dimensional filters, respectively, and two fully-connected layers that map the output of the last convolutional layer to the $18$ different class scores.
In designing our CNN architecture, we have aimed to use filters that are large-enough for learning patterns of tree textures and shapes.
We use a stride of $1$ in all convolutional layers to avoid information loss, and
keep the spatial dimensionality over convolutional layers via zero-padding.
While choosing the number of filters ($64$ filters per convolutional layer),
we have aimed to strike the right balance between having sufficient model
capacity and avoiding overfitting. We use max-pooling layers to achieve
partial translation invariance~\cite{Mishkin:2017}.
Finally, we have also investigated a number of similar deeper and wider architectures,
yet obtained the best performance with the presented network.
Additional details of the architecture\footnote{While Figure~\ref{fig:CNN}
shows an input with $3$ channels, the architecture can easily be adapted to any
number of input spectral bands. In general, for an input with $B$ bands, one
can simply use kernels of shape $5{\times}5{\times}B$ in the first layer.}
can be found in Figure~\ref{fig:CNN}.

We train the CNN model over the seen classes using cross-entropy loss, which corresponds to maximizing the label
log-likelihood in the training set.  To improve training, we employ Dropout regularization~\cite{Hinton:2012} (with
$0.9$ keep probability) and Batch Normalization~\cite{Ioffe:2015} throughout the network, excluding the last layer. Once
the network is trained, we use the output of the first fully connected layer,
i.e., the $128$-dimensional vector shown in Figure~\ref{fig:CNN},
as our image embedding $\phi(x)$ for the ZSL model. We additionally $\ell_2$-normalize this
vector, which is a common practice for CNN-based descriptors~\cite{razavian14cvpr}.

Finally, we note that one can consider pre-training the CNN model on external
large-scale data sets like ImageNet and fine-tuning it to the target problem.
While such an approach is likely to improve the recognition accuracy, it may
also lead to biased results due to potential overlaps between the classes in
our ZSL test set and the classes in the data set used during pre-training that
will violate the zero-shot assumption and will hinder the objectiveness of the
performance evaluation \cite{Xian:2017}. Therefore, we opt to train the CNN
model solely using our own training data set.

Additional CNN training details and an empirical comparison of our CNN model to
other contemporary classifiers are provided in Section \ref{sec:Experiments}.

\subsection{Class embedding}
\label{sec:class_embedding}

Class embeddings are the source of information for transferring knowledge that
is relevant to classification from seen to unseen classes. Therefore, the
embeddings need to capture the visual characteristics of the classes. For this
purpose, following the recent work on using multiple embeddings in computer vision
problems~\cite{Akata:2015}, we use a combination of three different
class embedding methods: (i) manually annotated attributes that we collect from
the target domain, (ii) text embeddings generated using unsupervised language
models, and, (iii) a hierarchical embedding based on a scientific taxonomy.

\begin{table}
\centering
\caption{Attributes for fine-grained tree categories \label{attribute_table}}
\footnotesize
\setlength{\tabcolsep}{2pt}
\begin{adjustbox}{width=\linewidth}
\begin{tabular}{@{}ll@{}} \toprule
Attribute type & Possible values\\ \midrule
    Height (feet)& \{10-15, 15-20, 20-25, 25-30, 30-40, 40-50, 50-60, 60-75\}\\
    Spread (feet) & \{10-15, 15-25, 25-35, 35-40, 40-50\}\\
    Crown uniformity & \{irregular outline, regular outline\}\\
    Crown density & \{open, moderate, dense\}\\
    Growth rate & \{medium, fast\}\\
    Texture & \{coarse, medium, fine\}\\
    Leaf arrangement & \{opposite/subopposite, alternate\}\\
    Leaf shape & \{ovate, star-shaped\}\\
    Leaf venation & \{palmate, pinnate\}\\
    Leaf blade length & \{0-2, 2-4, 4-8\}\\
    Leaf color & \{green, purple\}\\
    Fall color & \{green, yellow, purple, red, orange\}\\
    Fall characteristics & \{not showy, showy\}\\
    Flower color & \{brown, pink, green, red, white, yellow\}\\
    Flower characteristics & \{not showy, showy\}\\
    Fruit shape & \{round, elongated\}\\
    Fruit length & \{0-0.25, 0.25-0.50, 0.5-1.5, 1-3\}\\
    Fruit covering & \{dry-hard, fleshy\}\\
    Fruit color & \{brown, purple, green, red\}\\
    Fruit characteristics & \{not showy, showy\}\\
    Trunk bark branches & \{no thorns, thorns\}\\
    Pruning requirement & \{little, moderate\}\\
    Breakage & \{not resistant, resistant\}\\
    Light requirement & \{not part sun, part sun\}\\
    Drought tolerance & \{moderate, high\}\\ \bottomrule
\end{tabular}
\end{adjustbox}
\end{table}

Visual attributes are obtained by determining visually distinctive features of objects, such as their parts, textures, and shapes.
Since they provide a high-level description of object categories and their fine-grained properties, as perceived by
humans, attributes stand out as an outstanding class embedding method for zero-shot learning~\cite{Lampert:2014}. In order to utilize attributes in our work, we have collected $25$ attributes for tree species, based
on the Florida Trees Fact-Sheet~\cite{tree_attribute}. We list the names and possible values of these attributes in Table~\ref{attribute_table}.
These values are encoded as binary variables in a vector.

Although attributes provide powerful class embeddings, they are typically not comprehensive in capturing characteristics
of object categories, since attributes are defined in a manual way based on domain expertise.
Our second method that complements attributes is based on unsupervised word embedding models trained over large textual
corpora. For this purpose, we utilize the Word2Vec approach~\cite{Mikolov:2013}, which models the relationship between
words and their contexts. Since closely related words usually appear in similar contexts, the resulting word vectors are
known to implicitly encode semantic relationships. That is,
words with similar meanings typically correspond to nearby locations in the embedding space.  Our main goal here is to
leverage the semantic relationships encoded by Word2Vec to help the ZSL model in inferring models of unseen classes. For
this purpose, we use a $1000$-dimensional embedding model trained on Wikipedia articles, and extract word embeddings of
common names of tree species (given in Table \ref{data_partition_table}).
For categories with multiple words, we take the average of the per-word embedding vectors.

\begin{figure}
\centering
\includegraphics[width=\linewidth]{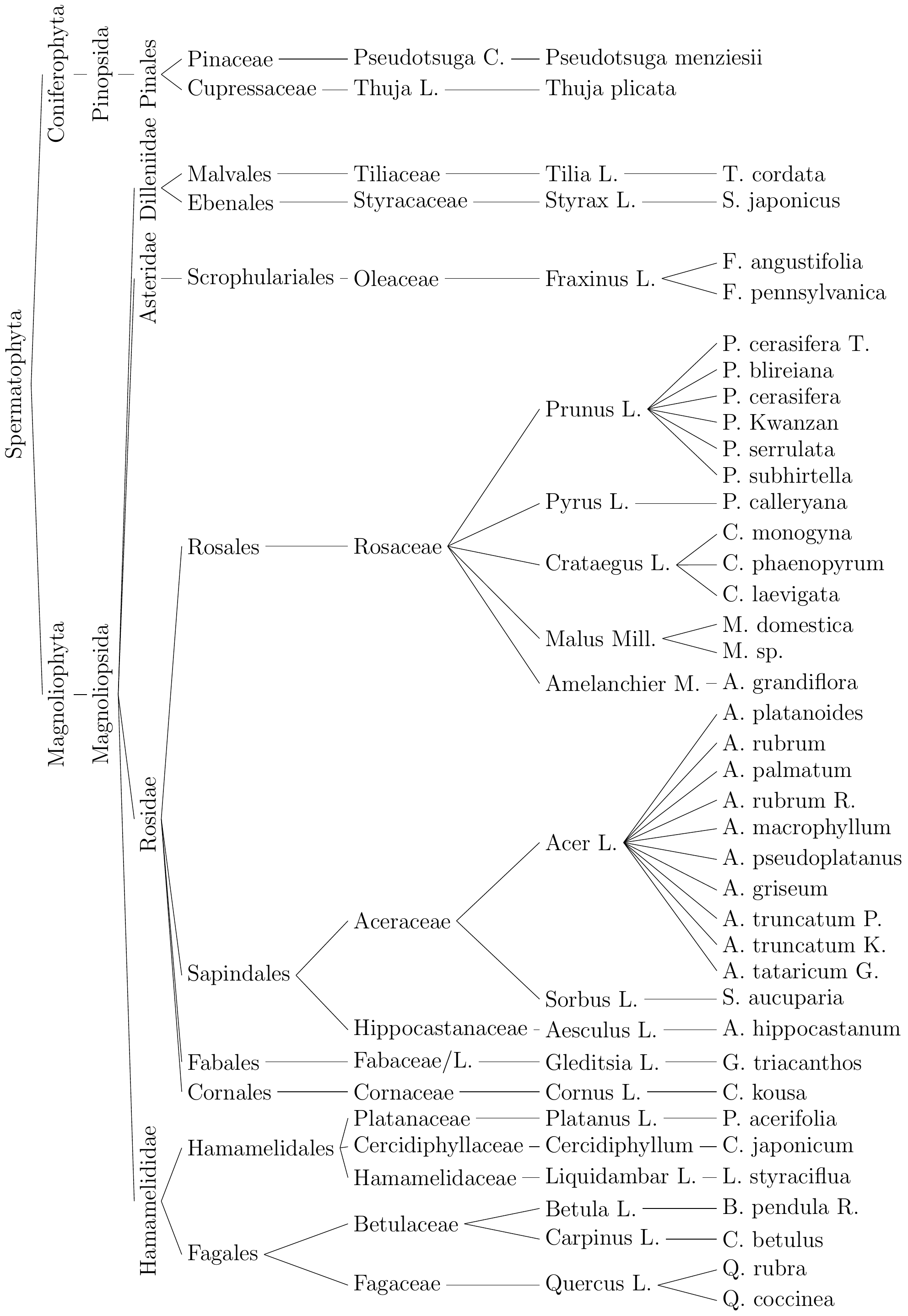}
\caption{Hierarchy embeddings are based on scientific classification of tree
species. This part of plant classification represents taxonomy of our tree
classes that starts with Spermatophyta superdivision and continues with the
names of division, class, subclass, order, family, genus, and species in order.
At each level, scientific names are written instead of common names.
Classification of each tree is taken from the Natural Resources Conservation
Service of the United States Department of Agriculture \cite{tree_hierarchy}.}
\label{classificationtaxonomy}
\end{figure}

The third and the last type of class embedding that we use aims to capture the similarities across tree species based on
their scientific classification. The scientific taxonomy of species in our data set is presented in
Figure~\ref{classificationtaxonomy}. Since the genetics of tree species directly affect their phenotype, the taxonomic
positions of trees can be informative about the visual similarity across the species.  In order to capture the position
and ancestors of tree species in the taxonomy tree, we apply the tree-to-vector conversion scheme described in
\cite{Mittelman:2014}. The embedding vector corresponding to a given tree species is obtained by defining a binary value
for each node in the taxonomy tree, and turning on only the values that correspond to the nodes that appear on the path
from the root to the leaf node of interest.  As a result, we obtain an embedding vector of length equivalent to the
number of nodes in the taxonomy.

We form the final embedding vector by concatenating the vectors produced by these three embedding methods.

\subsection{Joint bilinear and linear model}
\label{sec:joint_model}

The bilinear model specified in \eqref{compatibility_func} can be interpreted as learning a weighted
sum over all products of input and class embedding pairs. That is, the compatibility function can
be equivalently written in the following way:
\begin{equation}
    F(x,y) = \sum_{u=1}^d \sum_{v=1}^m W_{uv} [\phi(x)]_u [\psi(y)]_v
\end{equation}
where $[\phi(x)]_u$ and $[\psi(y)]_v$ denote the $u$-th and $v$-th dimensions of input and class embeddings,
respectively. From this interpretation we can see that the approach can learn relations between input and class
embeddings, but may not be able to evaluate the information provided by them individually.
To address
this shortcoming, we propose to extend the bilinear model by adding embedding-specific linear terms:
\begin{equation}
    F_e(x,y) = \phi(x)^\top W \psi(y) + w_x^\top \phi(x) + w_y^\top \psi(y) + b
    \label{extendedcompat}
\end{equation}
where $F_e$ is the {\em extended} compatibility function, $w_x$ is the linear model over the input embeddings, $w_y$ is
the linear model over the class embeddings, and $b$ is a bias term.

The advantage of having input and class embedding specific linear terms can be understood via the following examples: using the term $w_x^\top \phi(x)$, the model may adjust the entropy of the posterior probability distribution,
i.e., the confidence in predicting a particular class, by increasing or decreasing
all class scores depending on the clarity of object characteristics in the image.
Similarly, using the term $w_y^\top \psi(y)$, the model can estimate a class prior based on its embedding.
Finally, we note that the bias term has no effect on the estimated
class posteriors given by \eqref{softmax}, yet it simplifies the derivation below.

We incorporate the linear terms of the model in a practical way by simply
adding constant dimensions to both the input embedding and the class embedding. More specifically,
we extend the input and class embeddings as follows:
\begin{align}
    \phi_e(x) &= [\phi(x)^\top~1]^\top \\
    \psi_e(y) &= [\psi(y)^\top~1]^\top
\end{align}
where $\phi_e(x)$ and $\psi_e(y)$ denote the extended embedding vectors.
Similarly, we define the extended compatibility matrix $W_e$ as:
\begin{equation}
    W_e=\left[\begin{array}{cc}
            W  & w_{x}\\
            w_{y} & b
    \end{array}\right].
\end{equation}
It is easy to show that the bi-linear product $\phi_e(x)^\top W_e \psi_e(y)$
is equivalent to the extended compatibility function $F_e(x,y)$, given by \eqref{extendedcompat}. Therefore,
the linear terms can simply be introduced by adding bias dimensions to the embeddings.

\section{Experiments}
\label{sec:Experiments}

In this section, we present a detailed experimental analysis of our approach. We first describe our experimental setup. We
then present an evaluation of our CNN model in a supervised classification setting, followed by the evaluation of our
zero-shot learning approach. Finally, we experimentally analyze our model, compare it to important baselines, and discuss our
findings.

\subsection{Experimental setup}
\label{sec:expsetup}

In our experiments, we need to train and evaluate our approach in supervised and zero-shot learning settings. Therefore,
in order to obtain unbiased evaluation results, we need to define a principled way for tuning the model hyper-parameters.
This is particularly important in zero-shot learning because of the expectation
that the separation between the seen and unseen classes is clear. We follow the
guidance given in \cite{Xian:2017}: (i) ZSL should be evaluated mainly on least
populated classes as it is hard to obtain labeled data for fine-grained classes
of rare objects, (ii) hyper-parameters must be tuned on a validation class split
that is different training and test classes, and (iii) extracting image features
via a pre-trained deep neural network on a large data set should not involve
zero-shot classes for training the network.

Following these guidelines, we split the $40$ classes from our Seattle Trees
data set into three disjoint sets (with no class overlap): $18$ classes as the
{\em supervised-set}, $6$ classes as the {\em ZSL-validation} set, and the
remaining $16$ classes as the {\em ZSL-test} set.
The list of classes in each split is shown in Table~\ref{data_partition_table}.
We have arranged the splits roughly based on the number of examples in
each class: we mostly allocated the largest classes
to the supervised-set, the smallest classes to ZSL-validation, and the remaining ones to ZSL-test to have a reliable
performance for ZSL accuracy evaluation.

We use the supervised-set for two purposes: (i) to evaluate the CNN model in a supervised classification setting, and
(ii) to train the ZSL model using the supervised classes. For the supervised classification experiments, we use only the
classes inside the supervised-set, and we split the images belonging to these classes into {\em
supervised-train} ($60\%$), {\em supervised-validation} ($20\%$) and {\em supervised-test} ($20\%$) subsets. We
emphasize that these three subsets contain images belonging to the $18$ supervised-set classes, and they do not contain any
images belonging to a class from the ZSL-validation set or the ZSL-test set. We aim to maximize the performance on the
supervised-validation set when choosing the hyper-parameters of the supervised classifiers.

In ZSL experiments, we train the ZSL model using all images from the supervised-set. We use the zero-shot recognition
accuracy in the ZSL-validation set for tuning the hyper-parameters of the ZSL model. We evaluate the final model on
the ZSL-test set, which contains the unseen classes. In this manner, we avoid using unseen classes during training or
model selection, which, we believe, is fundamentally important for properly evaluating the ZSL models.

Throughout our experiments, we use normalized accuracy as the performance metric, which we obtain by averaging per-class
accuracy ratios. In this manner, we aim to avoid biases towards classes with a large number of examples.

\setlength{\tabcolsep}{0.05em}
\begin{table}[!t]
\caption{Class separation used for the data set and the number of instances in each class \label{data_partition_table}}
\centering
\begin{adjustbox}{width=\linewidth,center}
\begin{tabular}{|c c c c c c c c c c c c c c c c c c|c c c c c c|c c c c c c c c c c c c c c c c|}
\hline
\multicolumn{18}{|c|}{Supervised-set} &
\multicolumn{6}{c|}{\parbox[3cm][0.7cm][c]{1cm}{\centering ZSL-\\validation}} &
\multicolumn{16}{c|}{ZSL-test} \\ \hline
\rotatebox{270}{Midland Hawthorn (3154)} & \rotatebox{270}{Norway Maple (2970)} & \rotatebox{270}{Red Maple (2790)} & \rotatebox{270}{Cherry Plum (2510)} & \rotatebox{270}{Blireiana Plum (2464)} & \rotatebox{270}{Sweetgum (2435)} & \rotatebox{270}{Thundercloud Plum (2430)} & \rotatebox{270}{Kwanzan Cherry (2398)} & \rotatebox{270}{White Birch (1796)} & \rotatebox{270}{Littleleaf Linden (1626)} & \rotatebox{270}{Apple/Crabapple (1624)} & \rotatebox{270}{Red Oak (1429)} & \rotatebox{270}{Japanese Maple (1196)} & \rotatebox{270}{Red Maple (1086)} & \rotatebox{270}{Bigleaf Maple (885)} & \rotatebox{270}{Honey Locust (875)} & \rotatebox{270}{Western Red Cedar (720)} & \rotatebox{270}{Flame Ash (679)} & \rotatebox{270}{Chinese Cherry (1531)} & \rotatebox{270}{Washington Hawthorn (503)} & \rotatebox{270}{Paperbark Maple (467)} & \rotatebox{270}{Katsura (383)} & \rotatebox{270}{Norwegian Maple (372)} & \rotatebox{270}{Flame Amur Maple (242)} & \rotatebox{270}{London Plane (1477)} & \rotatebox{270}{Callery Pear (892)} & \rotatebox{270}{Horse Chestnut (818)} & \rotatebox{270}{Common Hawthorn (809)} & \rotatebox{270}{European Hornbeam (745)} & \rotatebox{270}{Sycamore Maple (742)} & \rotatebox{270}{Pacific Maple (716)} & \rotatebox{270}{Mountain Ash (672)} & \rotatebox{270}{Green Ash (660)} & \rotatebox{270}{Kousa Dogwood (642)} & \rotatebox{270}{Autumn Cherry (621)} & \rotatebox{270}{Douglas Fir (620)} & \rotatebox{270}{Orchard Apple (583)} & \rotatebox{270}{Apple Serviceberry (552)} & \rotatebox{270}{Scarlet Oak (489)} & \rotatebox{270}{Japanese Snowbell (460)} \\ \hline
\end{tabular}
\end{adjustbox}
\end{table}

\subsection{Supervised fine-grained classification}
\label{sec:evaluationfullysuper}

Before presenting our ZSL results, we first evaluate our CNN model in a supervised-setting to compare it against other
mainstream supervised classification techniques, and to give a sense of the difficulty of the fine-grained
classification problem that we propose. For this purpose, we use logistic regression and random forest
classifiers as our baselines. For a fair comparison, we train all methods on the supervised-train set, and tune their
hyper-parameters on the supervised-validation set.

We train our CNN architecture using stochastic gradient descent with the Adam
method~\cite{Adam:2014} that we also use for ZSL model estimation as described
in Section \ref{sec:compatibility}. Based on the
supervised-validation set, we have set the initial learning rate of Adam to $10^{-3}$, mini-batch size to $100$, and
$\ell_2$-regularization weight to $10^{-5}$. We also observed that it is beneficial to add perturbations of training
examples by randomly shifting each region with an amount in the range from zero to $20\%$ of the height/width.

We compare the resulting classifiers on the supervised-test set, as shown in Table~\ref{supervised_result}.  From these
results we can see that all classification methods perform clearly better than the random guess baseline ($5.6\%$). In
addition, we can see that the proposed CNN model both without perturbation ($27.9\%$) and with perturbation ($34.6\%$)
outperforms logistic regression ($16.4\%$) and random forest ($15.7\%$) by a large margin.

These results highlight the advantage of the deep image representation learned
by the CNN approach. In addition, we can observe the difficulty of the
fine-grained classification problem, which is quite different from the
traditional classification scenarios that aim to discriminate buildings from
trees or roads from grass. We believe that fine-grained classification is an
important open problem in remote sensing, and can lead to advances in object
recognition research.

\begin{table}[!t]
\renewcommand{\arraystretch}{1.0}
\caption{Supervised classification results (in \%) \label{supervised_result}}
\centering
\begin{tabular}{@{}lcc@{\hskip .1in}c@{\hskip .1in}c@{\hskip .1in}c}
\cmidrule[.8pt]{2-6}
\multicolumn{1}{l}{} &
\parbox{1.5cm}{\centering Random\\guess} & \parbox{1.5cm}{\centering Logistic\\regression} & \parbox{1.5cm}{\centering Random\\forest} & CNN & \parbox{1.5cm}{\centering CNN with\\perturbation} \\ \midrule
\parbox[4cm][0.6cm][c]{1.4cm}{\centering Normalized \\accuracy} & 5.6 & 16.4 & 15.7 & 27.9 & 34.6 \\ \bottomrule
\end{tabular}
\end{table}

\subsection{Fine-grained zero-shot learning}

In this section, we evaluate our ZSL approach and compare against three state-of-the-art ZSL methods:
ALE~\cite{Akata:2016}, SJE~\cite{Akata:2015}, and, ESZSL~\cite{Romera-Paredes:2015}. We train all ZSL models over the
supervised-train set, and tune all model hyper-parameters according to normalized accuracy on the ZSL-validation set.

For our approach, we initialize the $W$ matrix randomly from a uniform distribution~\cite{Glorot:2010}
and train the model using Adam optimizer~\cite{Adam:2014}. We tune the
hyper-parameters of initial learning rate of Adam and the number of training
iterations (for early-stopping based regularization).
For the ALE~\cite{Akata:2016} and SJE~\cite{Akata:2015} baselines, we use stochastic gradient descent (SGD) for training.
Unlike the original papers that use
a constant learning rate for SGD, we have found that decreasing the learning rate regularly over epochs
leads to better performance for these baselines. We tune the the learning rate policy on the ZSL-validation set.  For
the ESZSL~\cite{Romera-Paredes:2015} baseline, we tune its regularization
parameters $\lambda$ and $\gamma$ by choosing the
best-performing combination of the parameters in the range $\{10^{-3}, 10^{-2}, 10^{-1}, 10^{0}, 10^{1}, 10^{2},
10^{3}\}$ according to the ZSL-validation set and fix the $\beta$ hyper-parameter to $\lambda \gamma$, as suggested in
\cite{Romera-Paredes:2015}. In this case, the optimal compatibility matrix is given by a closed-form
solution~\cite{Romera-Paredes:2015}.
Finally, we note that all compared methods learn a single compatibility $W$ matrix, which provides a fair comparison across them.

For all methods, we have observed that imbalance in terms of the number of examples across the training classes can
negatively affect the resulting ZSL model. To alleviate this problem, we apply random over-sampling to the training set
such that the size of the training set for each class is equivalent to the size of the largest class.

\begin{table}[!t]
\renewcommand{\arraystretch}{0.9}
\caption{Zero-shot learning results (in \%) \label{tab:zsl_result}}
\centering
\begin{tabular}{@{}lcc@{\hskip .01in}c@{\hskip .01in}c@{\hskip .01in}c}
\cmidrule[.8pt]{2-6}
\multicolumn{1}{l}{} &
\parbox{1.5cm}{\centering Random\\guess} &
\parbox{1.4cm}{\centering ALE~\cite{Akata:2016}} &
\parbox{1.4cm}{\centering SJE~\cite{Akata:2015}} &
\parbox{1.4cm}{\centering ESZSL~\cite{Romera-Paredes:2015}} &
\parbox{1.4cm}{\centering Ours} \\ \midrule
\parbox[4cm][0.6cm][c]{1.4cm}{\centering Normalized \\accuracy} & $6.3$ & $12.5$ & $12.6$ & $13.2$ & $14.3$ \\ \bottomrule
\end{tabular}
\end{table}

The ZSL results over the $16$ ZSL-test classes are presented in Table \ref{tab:zsl_result}. Our ZSL model achieves a $14.3\%$
normalized accuracy, which is clearly better than the random guess baseline ($6.3\%$), ALE ($12.5\%$), SJE ($12.6\%$),
and ESZSL ($13.2\%$). These results validate the effectiveness of our probabilistic ZSL formulation.

The image embedding can have a profound effect on the ZSL performance. To better understand the efficacy of our
representation, we train our ZSL model over the outputs of different CNN layers (Figure~\ref{fig:CNN}), and tune the
number of training iterations on ZSL-validation for each one separately. When we use the $18$-dimensional classification
outputs, the ZSL performance drops from $14.3\%$ to $8.5\%$. Similarly, if we use the outputs of the layers
preceding the first fully-connected layer, the performance drops from $14.3$ to $13.0\%$ for last max-pooling output
(equivalently, the {\em Flatten} output), to $12.8\%$ for the convolutional layer with $3{\times}3$ kernels, and to
$11.1\%$ for the preceding max-pooling output. Simply using the original RGB image results in $8.3\%$ ZSL performance.
Overall, these results highlight the importance of the image representation on the ZSL performance, and suggest that the fully-connected
layer preceding the classification layer results in relatively generic features that are suitable for ZSL, in our architecture.

Our class embedding is a combination of three different embedding techniques.
To understand the contribution of each one, we present the ZSL performance for each possible combination of
class embedding methods in Table~\ref{zsl_embedding_result}. The first three rows of the table indicate that when the
embedding techniques are used individually, they result in a comparable performance, with a higher performance
for the Word2Vec ($12.1\%$), compared to attributes ($8.4\%$) and hierarchy ($9.7\%$). The following three rows indicate
that the hierarchy-Word2Vec ($13.2\%$) embedding pair leads to better results compared to individual
embeddings as well as the pairs of attribute-Word2Vec ($12.6\%$) or hierarchy-attribute ($11.2\%$).
These results show that our hierarchy and Word2Vec embeddings are more effective than attribute embeddings. This observation suggests
that the
recognition accuracy can be improved possibly by defining more descriptive attributes.
On the other hand, the final result based on the combination of all embeddings, which leads to the highest accuracy
($14.3\%$), shows that our class embeddings are complementary to each other.

Another important aspect of the proposed method is extending the bilinear model by adding linear terms for the input
and class embeddings. To understand the significance of this extension, we present
an evaluation of the linear terms in Table~\ref{zsl_linearterm_result}. The table shows that without having any
linear term, the normalized accuracy for ZSL is $11.8\%$. Adding $w_x^\top\phi(x)$, see \eqref{extendedcompat}, improves
the performance to $12.2\%$, and adding $w_y^\top\psi(y)$ improves the performance to $13.4\%$.
Finally, adding both terms together leads to our highest result of $14.3\%$ normalized accuracy. These
results validate the importance of adding linear terms into the bilinear ZSL model.

\subsection{Analysis and discussion}

The results presented so far show that the proposed ZSL approach performs
significantly better than the random guess baseline, and
also better than several other state-of-the-art ZSL methods. However, an important question is how well
ZSL performs in a practical sense. To address this question, we compare our ZSL approach against supervised
classification, and discuss the relative advantages and disadvantages of supervised versus zero-shot learning of
novel class models.

\begin{table}[!t]
\renewcommand{\arraystretch}{1.2}
\caption{Effect of different class embeddings on zero-shot
learning performance (in \%) \label{zsl_embedding_result}}
\centering
\def\mycolskip{\hskip .1in}
\begin{tabular}{@{}c@{\mycolskip}c@{\mycolskip}c@{\mycolskip}c@{}} \toprule
Attribute & Hierarchy & Word2Vec & \parbox{1.5cm}{\centering Normalized \\accuracy }\\ \midrule
\cmark & \textcolor{gray}{\xmark} & \textcolor{gray}{\xmark} & 8.4 \\ 
\textcolor{gray}{\xmark} & \cmark & \textcolor{gray}{\xmark} & 9.7 \\
\textcolor{gray}{\xmark} & \textcolor{gray}{\xmark} & \cmark & 12.1 \\
\cmark & \cmark & \textcolor{gray}{\xmark} & 11.2 \\
\cmark & \textcolor{gray}{\xmark} & \cmark & 12.6 \\
\textcolor{gray}{\xmark} & \cmark & \cmark & 13.2 \\
\cmark & \cmark & \cmark & {\bf 14.3} \\ \bottomrule
\end{tabular}
\end{table}

\begin{table}[!t]
\renewcommand{\arraystretch}{1.2}
\caption{Effect of linear terms on zero-shot performance
(in \%) \label{zsl_linearterm_result}}
\centering
\def\mycolskip{\hskip .1in}
\begin{tabular}{@{}c@{\mycolskip}c@{\mycolskip}c@{\mycolskip}c@{}} \toprule
\parbox{1.5cm}{\centering For image embedding} & \parbox{1.5cm}{\centering For class embedding } & \parbox{1.5cm}{\centering Normalized \\accuracy }\\ \midrule
\textcolor{gray}{\xmark} & \textcolor{gray}{\xmark} & 11.8 \\
\cmark & \textcolor{gray}{\xmark} & 12.2 \\
\textcolor{gray}{\xmark} & \cmark & 13.4 \\
\cmark & \cmark &  {\bf 14.3} \\ \bottomrule
\end{tabular}
\end{table}

For this purpose, we use five-fold cross validation over the whole ZSL-test set, where repeatedly one of the folds is
utilized for training the supervised classifiers, and the remaining folds are utilized as the test subset.
In our analysis, we consider two types of supervised classifiers: (i) CNN models that are trained from scratch over
supervised ZSL-test examples only, (ii) pre-trained CNN models that are {\em fine-tuned} to the ZSL-test classes. For
the latter approach, we re-initialize and re-train the last layer, i.e., the classification layer, of our pre-trained CNN model.
Our motivation for fine-tuning is that all layers preceding the last layer are likely to extract a class-agnostic
image representation, and the last layer can be interpreted as a linear classifier that transforms the learned image
representation into the classification scores.  In this way, we can effectively transfer knowledge from supervised-set
to ZSL-test, using supervised training examples for the latter set.

It is well-known that the accuracy of a supervised classifier tends to improve as its training set gets larger. In this context,
to understand the trade-off between using a ZSL approach, which uses zero training examples for the target classes, versus collecting supervised training examples, we train the supervised classifiers at varying number of training examples. More specifically,
we train separate supervised classifiers by limiting the number of examples {\em per class} to each possible constant in $1,2,4,\dots,2^{10}$.
We impose these limits by subsampling the training subset at each fold of five-fold cross-validation.
To obtain reliable statistics, we repeat each experiment $10$ times.

\begin{figure}[!t]
\centering
\includegraphics[width=\linewidth]{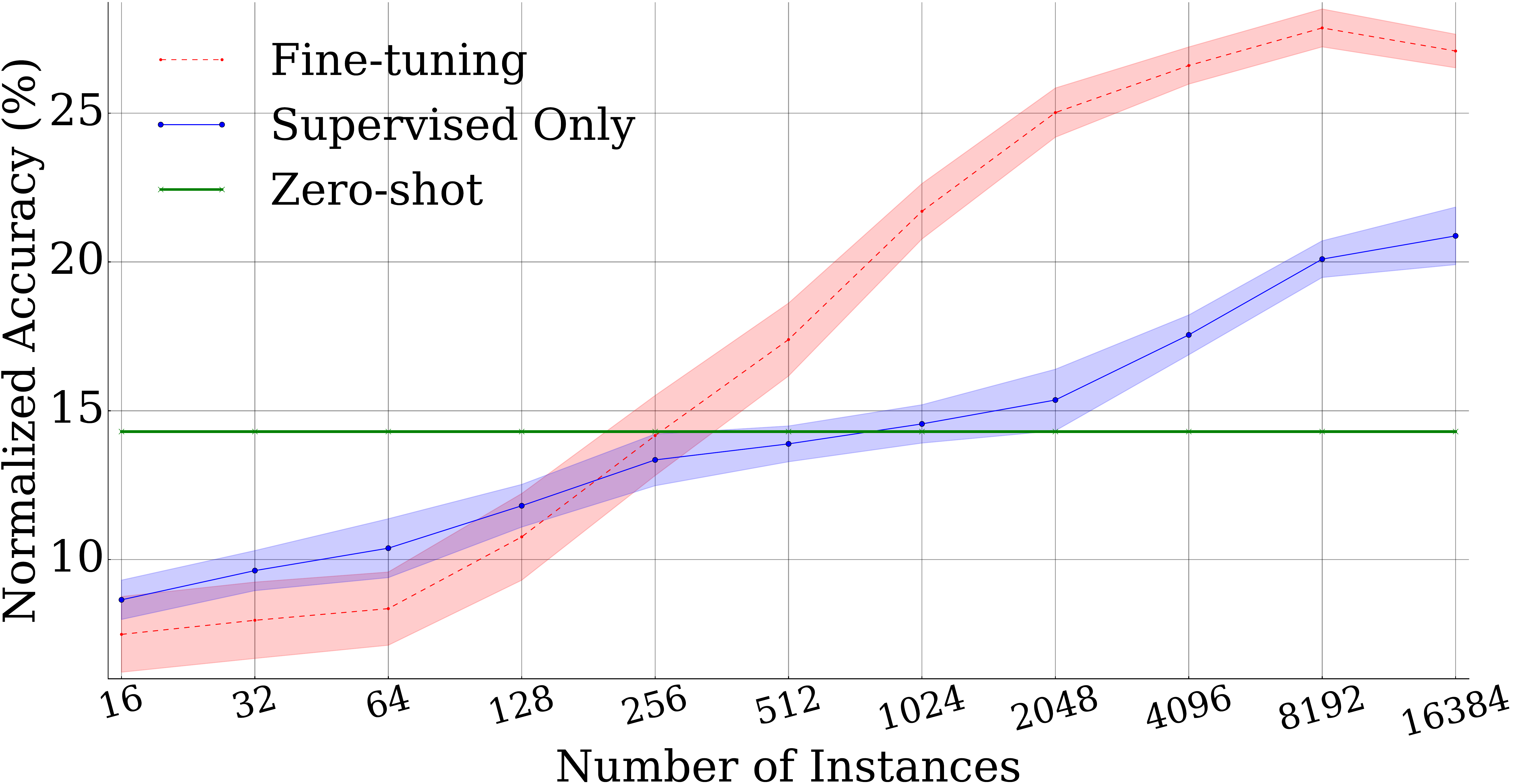}
\caption{Performance comparison of the proposed framework with fine-tuning and
supervised-only methods on zero-shot test classes. Fine-tuning and
supervised-only results are derived at different points when the number of
instances is increasing. The $x$-axis is shown in log-scale.}
\label{upper_bound}
\end{figure}

Figure~\ref{upper_bound} presents the results for the supervised-only CNN,
fine-tuned CNN, and the ZSL model. The $x$-axis shows the number of training
examples for the supervised classifiers, and the $y$-axis shows the corresponding normalized
accuracy scores. The curves are obtained by averaging results over all folds and all runs, and setting the curve thickness
to the standard deviation of the results. The ZSL approach is shown as a single horizontal line, as it does not use any supervised
training examples.

From the results we can see that supervised-only CNN starts to match the ZSL performance only when the number of training examples is
more than $512$, and the fine-tuned CNN reaches the ZSL performance at $256$ samples. This is a significant achievement considering that
(i) ZSL approach uses zero-training examples from the target classes, and (ii)
we are working with fine-grained categories that are hard to distinguish even
by visual inspection of the image data. We expect ZSL performance
to further improve following the advances in image representation, image
resolution, class embeddings, and ZSL formulations.

\begin{figure}[!t]
\centering
\includegraphics[width=.7\linewidth]{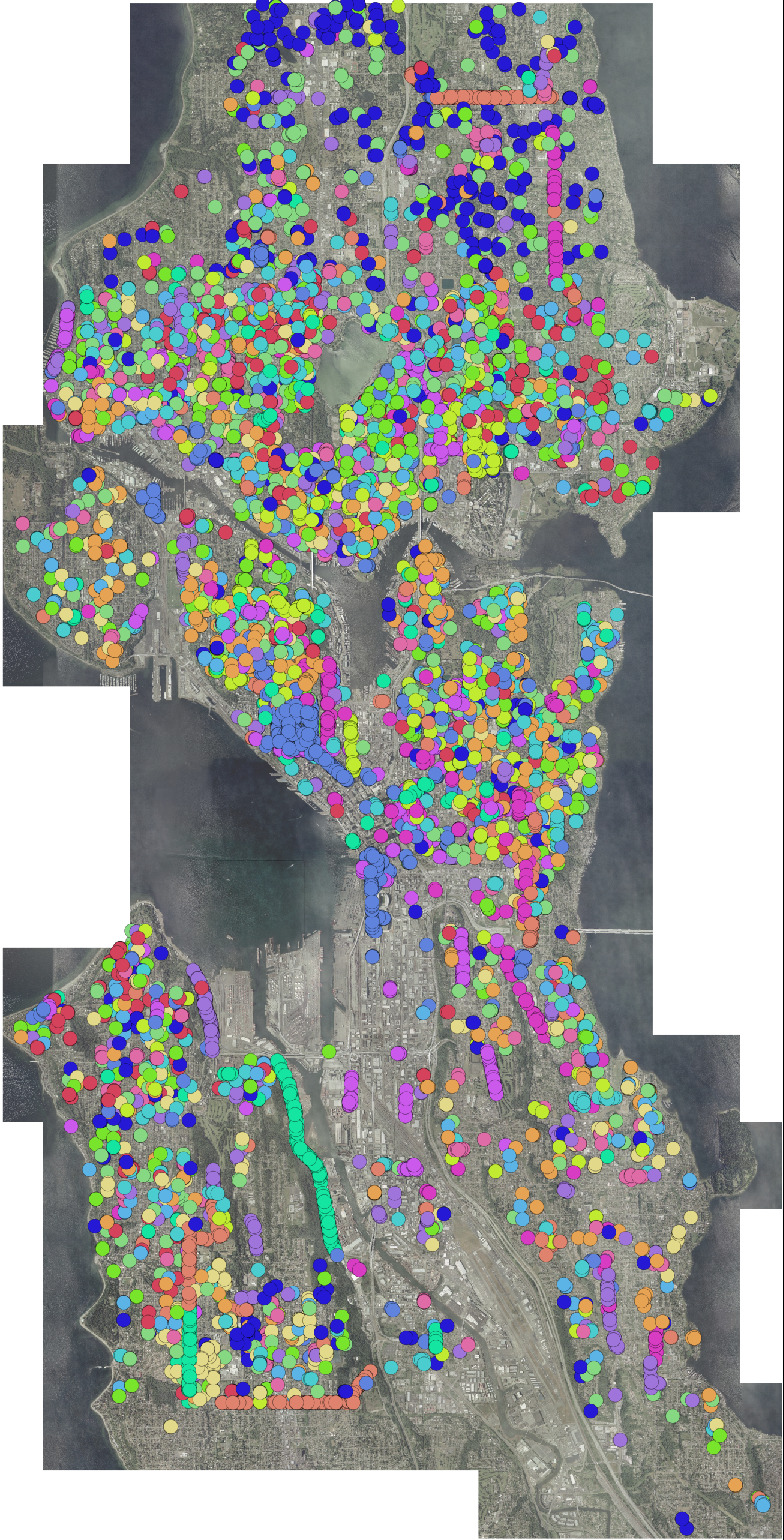}
\caption{Spatial distribution of instances belonging to the zero-shot test
(unseen) classes. Each point shows one instance, and the point colors represent
the classes. (Best viewed in color.)}
\label{travel_area_by_point}
\end{figure}

Importantly, we should also note that the collection and annotation of even
$256$ training examples can be a very costly task: sample collection may
require spatially surveying a very large area, and annotating them with class labels typically requires
inspection of the instances or their close-by pictures by domain-experts, as the most fine-grained categories are very
difficult to distinguish. For example, Figure~\ref{travel_area_by_point} illustrates the Seattle region and the
spatial distribution of our $16$-class ZSL-test instances in this region. In this figure, we observe that the instances and
classes are scattered all around an area of $217~\textrm{km}^2$, which casts the
data collection and annotation a very time-consuming and challenging task. In this context, we believe that zero-shot
learning of fine-grained categories can potentially become a central topic towards building semantically rich
image understanding systems for remote sensing.

\section{Conclusions}
\label{sec:Conclusions}

We studied the zero-shot learning problem for fine-grained object recognition
in remotely sensed images. To cope with the difficulty of learning a
very large number of very similar object categories as well as the need for being
able to recognize classes even when there are no training examples, our framework
exploited alternative sources of auxiliary information to build an association
between the seen and unseen classes. The proposed approach learned a bilinear
function from the seen classes so that the compatibility between the visual characteristics
observed in the input image data and the auxiliary information that described
the semantics of the classes of interest is modeled. Then, we showed how this compatibility
function could be used for performing knowledge transfer during the inference
of the unseen classes. Extensive experiments using different partitionings of a
challenging aerial data set with $40$ types of street trees defined as fine-grained
target classes showed that our method obtained $14.3\%$ classification accuracy,
which was significantly better than random guessing ($6.3\%$) for $16$ test
classes and three other zero-shot learning algorithms from the literature.
Future work includes new representations for auxiliary information that
models different aspects of spectral and spatial data characteristics as well
as the domain-specific class semantics.

\bibliographystyle{IEEEtran}

\begin{IEEEbiography}[{\includegraphics[width=1in,height=1.25in,clip,keepaspectratio]{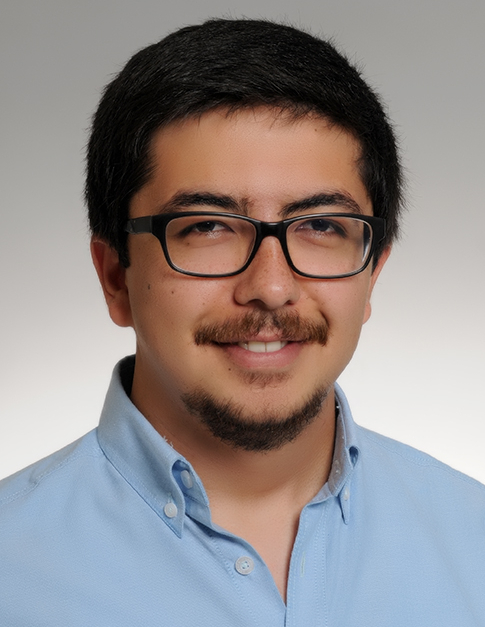}}]{Gencer Sumbul}
    received the B.S. degree in Computer Engineering from Bilkent University,
    Ankara, Turkey, in 2015.
    He is currently pursuing the M.S. degree
    in Computer Engineering at Bilkent University. His
    research interests include computer vision, deep learning and machine
    learning, with special interest in fine-grained object recognition with weak supervision on remote sensing images.
\end{IEEEbiography}

\begin{IEEEbiography}[{\includegraphics[width=1in,height=1.25in,clip,keepaspectratio]{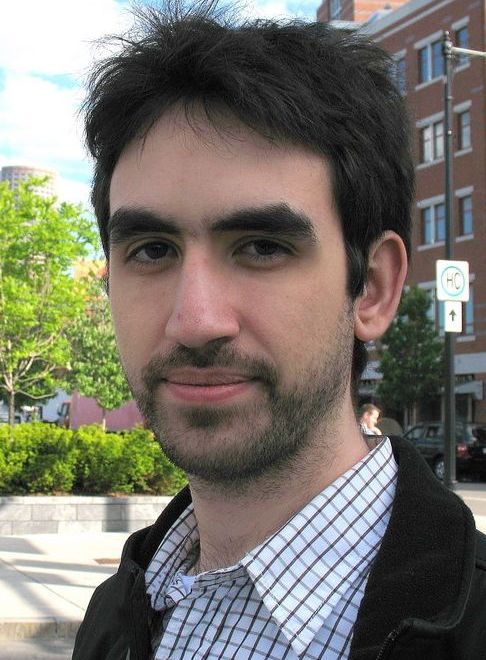}}]{Ramazan Gokberk Cinbis}
    graduated from Bilkent University, Turkey, in 2008, and
    received an M.A. degree from Boston University, USA, in 2010.
    He was
    a doctoral student in the LEAR team, at INRIA Grenoble,
    France, between 2010-2014, and received a PhD degree in
    computer science from Universit\'{e} de Grenoble,
    in 2014. He is currently an assistant professor at Middle East Technical University.
    His research areas include computer vision and machine learning,
    with special interest in statistical image models, and deep learning with weak supervision.
\end{IEEEbiography}

\begin{IEEEbiography}[{\includegraphics[width=1in,height=1.25in,clip,keepaspectratio]{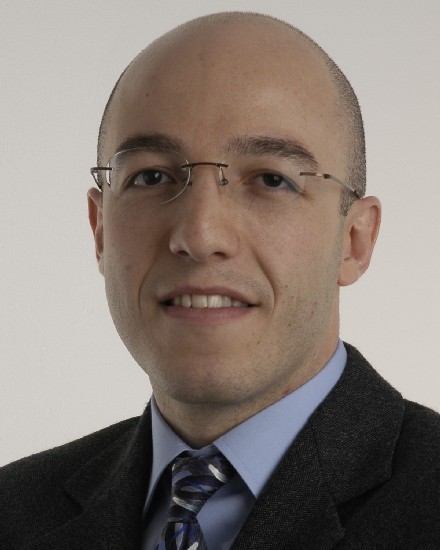}}]{Selim Aksoy}
(S'96-M'01-SM'11) received the B.S.\ degree from the Middle East Technical University,
Ankara, Turkey, in 1996 and the M.S.\ and Ph.D.\ degrees from the University
of Washington, Seattle, in 1998 and 2001, respectively.

He has been working at the Department of Computer Engineering, Bilkent
University, Ankara, since 2004, where he is currently an Associate Professor.
He spent 2013 as a Visiting Associate Professor at the Department of Computer
Science \& Engineering, University of Washington.
During 2001--2003, he was a Research Scientist at Insightful
Corporation, Seattle, where he was involved in image understanding and data
mining research sponsored by the National Aeronautics and Space Administration,
the U.S.\ Army, and the National Institutes of Health.
His research interests include computer vision, statistical and structural pattern
recognition, machine learning and data mining with applications to remote
sensing, medical imaging, and multimedia data analysis.

Dr.~Aksoy is a member of the IEEE Geoscience and Remote Sensing Society, the
IEEE Computer Society, and the International Association for Pattern Recognition
(IAPR).
He was one of the Guest Editors of the special issues on
Pattern Recognition in Remote Sensing of IEEE Transactions on Geoscience and
Remote Sensing, Pattern Recognition Letters, and IEEE Journal of Selected Topics
in Applied Earth Observations and Remote Sensing in 2007, 2009, and 2012,
respectively. He served as the Vice Chair of the IAPR Technical Committee 7 on
Remote Sensing during 2004--2006, and as the Chair of the same committee during
2006--2010. He also served as an Associate Editor of Pattern Recognition Letters
during 2009--2013.
\end{IEEEbiography}

\end{document}